\documentclass{llncs}

\usepackage{eccv}
\usepackage{eccvabbrv}
\usepackage{graphicx}
\usepackage{booktabs}
\usepackage{array}
\usepackage{xcolor}
\usepackage{xspace}
\usepackage{tikz}
\usetikzlibrary{positioning}
\usepackage[accsupp]{axessibility}
\usepackage[pagebackref,breaklinks,colorlinks,citecolor=eccvblue]{hyperref}

\newcommand{\phasemarg}{Phase Marginalization\xspace}
\newcolumntype{L}[1]{>{\raggedright\arraybackslash}p{#1}}
\newcommand{\gtacity}{GTA5$\rightarrow$\allowbreak Cityscapes}
\newcommand{\synthiacity}{SYNTHIA$\rightarrow$\allowbreak Cityscapes}
\newcommand{\dinovtwofstable}{DINOv2/F-\allowbreak stable}
\newcommand{\learnedphaseattn}{Learned Phase-Feature Attention Aggregation\xspace}
\newcommand{\partialphaseattn}{Partial Encoder Tuning with Phase-Feature Attention\xspace}
\newcommand{\partialspatialphaseattn}{Partial Encoder Tuning with Spatial Phase Attention\xspace}
\newcommand{\boundaryphaseconsistency}{Boundary-Aware Phase Consistency Aggregation\xspace}
\newcommand{\localvarianceweighting}{Deterministic Local-Variance Reliability Weighting\xspace}
\newcommand{\learnedreliabilityweighting}{Learned Non-Competitive Reliability Weighting\xspace}
\newcommand{\pretransformerembedavg}{Pre-Transformer Patch-Embedding Averaging\xspace}
\newcommand{\depthgatedfusion}{Depth-Specific Confidence-Gated Residual Phase Fusion\xspace}

\begin{document}

\title{Phase Marginalization for Patch-Grid Instability in Vision Transformers}
\author{Oğuzhan Ercan\inst{1}}
\institute{Scientific and Technological Research Council of Türkiye\\
\email{oguzhanercancs@gmail.com}\\
}

\maketitle
\pagestyle{plain}

\begin{abstract}
Vision Transformers operate on fixed patch grids, which can introduce phase-dependent instability for dense prediction: changing the patch partition can change the token evidence available to a pixel, especially near boundaries. We formalize patch-grid phase as a nuisance variable and propose Phase Marginalization, a post-hoc marginalization method that evaluates structured patch-grid phases, inverse-aligns dense outputs, and aggregates them in the original image coordinate system. The central variant, Uniform \phasemarg with $K=4$, is training-free and improves over the canonical $K=1$ baseline across measured segmentation, depth, and local matching settings. In a controlled Cityscapes experiment, Uniform \phasemarg provides a modest compute-matched advantage over generic shift-based four-forward test-time augmentation (TTA) (+0.31 mean Intersection-over-Union over the strongest tested generic row). A scaling study further shows that $K=4$ is a practical cost-accuracy trade-off: $K=8$ is essentially unchanged and $K=16$ adds little accuracy at much higher latency. These results position patch-grid phase as a measurable nuisance variable and \phasemarg as a simple diagnostic and post-hoc marginalization baseline for dense ViT prediction.
\keywords{Vision Transformers \and Dense prediction \and Test-time augmentation \and Robustness}
\end{abstract}

\section{Introduction}
\label{sec:intro}

Dense prediction tasks require coordinate-stable outputs: a pixel-level label should not depend sensitively on arbitrary details of how an image is partitioned before inference. Vision Transformers (ViTs)~\cite{dosovitskiy2020}, however, convert an image into fixed-size patch tokens. This patchification step introduces a discrete spatial phase: the same scene can be represented by different token memberships when the patch grid is shifted by a few pixels. Near semantic boundaries, where neighboring pixels often belong to different classes or depths, this phase can change the evidence available to the dense prediction head.

We study this patch-grid phase as a nuisance variable. For a patch size $P$, a phase offset $\phi=(d_x,d_y)$ specifies where the patch grid begins relative to the image coordinate system. If dense predictions produced under different sub-patch phases disagree after being mapped back to the same pixel coordinates, the model exhibits patch-grid phase instability. This perspective separates a specific architectural quantization effect from generic image perturbations.

\phasemarg addresses this nuisance variable directly. Instead of redesigning or retraining the backbone, the uniform method evaluates a frozen ViT and dense head under a small set of patch-size-defined phases, inverse-aligns the phase-specific dense outputs, and averages them. The method is therefore a structured, post-hoc marginalization procedure. It is compatible with frozen DINO-style backbones~\cite{oquab2023,simeoni2025dinov3} and can be implemented as an inference wrapper around an existing dense prediction model.

The method is related to test-time augmentation (TTA), but it samples a different object. Generic TTA averages predictions over image transformations such as flips, crops, or shifts. \phasemarg samples offsets of the patch grid itself and uses exact inverse alignment before aggregation. Our compute-matched Cityscapes comparison tests this distinction under the same four-forward budget and finds a modest positive margin for structured phase sampling.

Architectural methods such as DPT~\cite{ranftl2021dpt}, ViT-Adapter~\cite{chen2022vitadapter}, Swin~\cite{liu2021swin}, PVT~\cite{wang2021pvt}, and overlapping-token designs~\cite{xiao2021early} address dense prediction or spatial structure by changing the model or training regime. These methods answer different deployment questions: architectural redesign, adapter training, or alternative backbone construction. Our controlled empirical question is narrower: for a fixed dense predictor, does marginalizing the patch-grid phase improve over canonical inference and the tested generic shift-based TTA controls?

Our contributions are:
\begin{itemize}
    \item \textbf{Phenomenon.} We define patch-grid phase instability as a measurable nuisance variable for dense ViT prediction.
    \item \textbf{Method.} We propose Uniform \phasemarg, a training-free post-hoc marginalization over structured patch-grid phases with inverse-aligned dense outputs.
\item \textbf{Evidence.} We show that Uniform $K=4$ improves over $K=1$ across measured segmentation, depth, and local matching settings and demonstrate cross-domain generalisation by evaluating models trained on synthetic datasets (GTA5, SYNTHIA) and tested on real images (Cityscapes).
    \item \textbf{Controls.} We report compute-matched Cityscapes comparisons against generic shift-based TTA and a Cityscapes $K$/latency scaling study that motivates $K=4$ as a practical default.
\end{itemize}

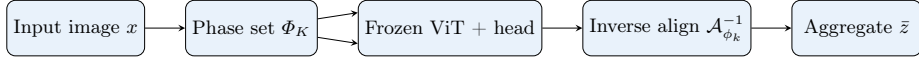
\begin{figure*}[t]
    \centering
    \resizebox{\textwidth}{!}{%
    \begin{tikzpicture}[node distance=0.65cm,>=stealth,every node/.style={transform shape}]
    \node (input) [draw, rounded corners, fill=eccvblue!10, minimum width=1.8cm, minimum height=0.9cm] {Input image $x$};
    \node (phase) [draw, rounded corners, fill=eccvblue!10, minimum width=2.0cm, minimum height=0.9cm, right=of input] {Phase set $\Phi_K$};
    \node (vit) [draw, rounded corners, fill=eccvblue!10, minimum width=2.4cm, minimum height=0.9cm, right=of phase] {Frozen ViT + head};
    \node (align) [draw, rounded corners, fill=eccvblue!10, minimum width=2.7cm, minimum height=0.9cm, right=of vit] {Inverse align $\mathcal{A}^{-1}_{\phi_k}$};
    \node (avg) [draw, rounded corners, fill=eccvblue!10, minimum width=2.1cm, minimum height=0.9cm, right=of align] {Aggregate $\bar{z}$};
    \draw[->] (input) -- (phase);
    \draw[->] ([yshift=0.15cm]phase.east) -- ([yshift=0.25cm]vit.west);
    \draw[->] ([yshift=-0.15cm]phase.east) -- ([yshift=-0.25cm]vit.west);
    \draw[->] (vit) -- (align);
    \draw[->] (align) -- (avg);
    \end{tikzpicture}
    }
    \caption{Conceptual schematic of Phase Marginalization. A single frozen dense predictor is evaluated under structured patch-grid phase offsets; each output is inverse-aligned to the original image coordinate system and then aggregated by averaging logits.}
    \label{fig:method-schematic}
\end{figure*}

\section{Related Work}
\label{sec:related}

\noindent\textbf{Translation equivariance and aliasing.}
Small translations can expose aliasing in visual recognition systems. CNNs are not automatically shift-invariant when downsampling is present, and anti-aliasing filters can improve shift consistency~\cite{azulay2018,zhang2019}. Our focus is a distinct source of aliasing: the non-overlapping patch grid used to tokenize ViT inputs.

\noindent\textbf{ViT patchification and tokenization artifacts.}
ViTs tokenize images into non-overlapping patches~\cite{dosovitskiy2020}. Several designs reduce local artifacts by modifying the tokenizer or early feature extractor, including early convolutions~\cite{xiao2021early}, conditional or locally structured encodings~\cite{chu2021twins}, shifted windows~\cite{liu2021swin}, and pyramid/overlapping token designs~\cite{wang2021pvt}. \phasemarg takes a complementary route: it keeps the model fixed and marginalizes over a discrete set of patch-grid phases at inference time.

\noindent\textbf{Test-time augmentation and ensembling.}
TTA averages predictions across transformations and is widely used as an inference-time ensemble~\cite{shorten2019,touvron2021}. Test-time adaptation methods such as TENT~\cite{wang2022tent} modify model parameters or statistics at test time. Uniform \phasemarg has the same broad cost profile as a $K$-forward TTA ensemble, but the sampled transformations are patch-size-defined phases with inverse alignment, not arbitrary semantic or photometric augmentations.

\noindent\textbf{Dense prediction architectures.}
DPT~\cite{ranftl2021dpt} and ViT-Adapter~\cite{chen2022vitadapter} adapt ViTs for dense prediction using dedicated decoders or adapter modules. Swin and PVT-style architectures introduce hierarchical or windowed inductive biases~\cite{liu2021swin,wang2021pvt}. These methods change the architecture or training recipe, whereas Uniform \phasemarg is a post-hoc inference wrapper. They therefore serve as related deployment families rather than fixed-predictor inference comparisons.

\noindent\textbf{Denoising and feature-artifact mitigation.}
Denoising ViT-style methods study artifacts in ViT feature maps and mitigate them through feature denoising or learned correction mechanisms. This is adjacent to our motivation, but the intervention is different: \phasemarg does not denoise features or train a correction module for the backbone. It treats the patch-grid origin as an explicit phase variable and marginalizes dense predictions over structured phase offsets at inference time. We therefore discuss DVT-style methods as related artifact-mitigation work rather than compute-matched baselines.

\section{Method}
\label{sec:method}

\subsection{Patch-grid phase instability}

Let $x\in\mathbb{R}^{3\times H\times W}$ be an input image and let $P$ be the patch size. A patch-grid phase is an offset
\begin{equation}
\phi=(d_x,d_y), \qquad d_x,d_y\in\{0,\ldots,P-1\}
\end{equation}.
For a dense prediction model with frozen encoder $f_\theta$ and head $h$, the canonical single-phase baseline is $\phi=(0,0)$, denoted $K=1$.

We define a phase-shift operator $\mathcal{S}_\phi$ that changes the patch-grid offset using reflective padding and cropping. For phase $\phi$, the model produces a dense logit map
\begin{equation}
 z_\phi = h(f_\theta(\mathcal{S}_\phi(x)))
\end{equation}.
The output $z_\phi$ is expressed in the shifted padded coordinate system. Before comparing or aggregating phases, it is mapped back to the original image coordinates by an inverse-alignment operator $\mathcal{A}^{-1}_\phi$: patch outputs are reshaped to the raster patch grid, bilinearly upsampled to the padded image resolution with align-corners set to false, and cropped by the inverse phase offset to recover the original $H\times W$ field.

For aligned phase features $F_{\phi_k}$, phase instability can be summarized by per-pixel phase variance,
\begin{equation}
\sigma^2(p)=\frac{1}{K}\sum_{k=1}^K \lVert F_{\phi_k}(p)-\bar{F}(p)\rVert_2^2,
\qquad
\bar{F}(p)=\frac{1}{K}\sum_{k=1}^K F_{\phi_k}(p)
\label{eq:phase-var}
\end{equation}.
Lower phase variance indicates less dispersion among phase-conditioned representations at the same pixel or descriptor location. Equation~\ref{eq:phase-var} defines this diagnostic; Section~\ref{sec:boundary-phase} uses it to measure phase-conditioned descriptor variability, while task metrics evaluate whether marginalization improves predictions.

\subsection{Uniform Phase Marginalization}

Given a discrete phase set $\Phi_K=\{\phi_1,\ldots,\phi_K\}$, Uniform \phasemarg averages inverse-aligned logits:
\begin{equation}
\bar{z}(p)=\frac{1}{K}\sum_{k=1}^K \mathcal{A}^{-1}_{\phi_k}(z_{\phi_k})(p)
\label{eq:uniform}
\end{equation}.
Equation~\ref{eq:uniform} defines the uniform marginalization of logits. This is the central method in the paper. It does not update the encoder, dense head, or normalization statistics. For $K=4$ and even patch size $P$, we use
\begin{equation}
\Phi_4=\{(0,0),(0,P/2),(P/2,0),(P/2,P/2)\}
\end{equation}.
This samples four canonical quadrants of the sub-patch phase space. Section~\ref{sec:k-scaling} evaluates the cost-accuracy trade-off of this choice on Cityscapes.

\subsection{Secondary learned and adapted variants}

We evaluate learned and adapted variants as a secondary audit, not as the main method. \learnedphaseattn is a trained feature-level variant, distinct from Uniform \phasemarg. For each input image, $K$ phase-shifted views are evaluated with reflect padding. Patch tokens from each phase are reshaped to the raster patch grid, bilinearly upsampled with align-corners set to false, cropped back to the original coordinate frame, and optionally downsampled to a working stride $s=4$ for memory efficiency. This produces an aligned phase feature stack $F\in\mathbb{R}^{H/s\times W/s\times K\times D}$.

At each spatial location, the model computes phase-feature variance and concatenates, for each phase $k$, the phase feature $F_k(p)$, the phase-variance vector, and a learned phase embedding $e_k\in\mathbb{R}^{16}$. For $D=768$, the attention input has dimension $2D+16=1552$. A shared multilayer perceptron (MLP) with architecture $1552\rightarrow1024\rightarrow256\rightarrow1$ and GELU activations produces one score per phase. The scores are softmax-normalized over phases at each spatial location. For aligned features $F_{\phi_k}$, this gives
\begin{equation}
\alpha_k(p)=\frac{\exp s_k(p)}{\sum_j\exp s_j(p)},
\qquad
\tilde{F}(p)=\sum_{k=1}^K\alpha_k(p)F_{\phi_k}(p)
\end{equation},
followed by LayerNorm and the frozen dense head. In the segmentation implementation, only the phase-attention MLP is trained; the encoder and segmentation head remain frozen.

\partialphaseattn uses the same aligned $K$-phase feature stack and learned phase attention, but also unfreezes the last encoder blocks and layer-normalization parameters while keeping the segmentation head frozen. \partialspatialphaseattn replaces the larger phase-attention MLP with a spatial per-pixel phase-attention module using architecture $2D+16\rightarrow512\rightarrow128\rightarrow1$, with scores softmax-normalized over phases independently at each spatial location. \pretransformerembedavg is a negative design variant: it averages patch embeddings from multiple phase-shifted inputs before transformer self-attention and then runs the transformer once on the averaged tokens. \depthgatedfusion is a separate depth-estimation variant with residual correction, confidence gating, calibration, and geometric/depth losses; it should not be conflated with either Uniform \phasemarg or the segmentation learned phase-attention aggregator.

Other auxiliary audits use the same descriptive convention. \boundaryphaseconsistency adds a boundary-band consistency loss that pulls the aggregate toward the most confident phase feature and away from the least confident one near class transitions. Reliability-weighted variants such as \localvarianceweighting and \learnedreliabilityweighting change only how phase weights are estimated; they are secondary reliability probes rather than the central training-free method.

\subsection{Relation to generic TTA}

Generic TTA samples image transformations and averages inverse-transformed predictions. \phasemarg samples the patch-grid phase variable induced by tokenization. The methods can have the same number of forward passes, but they differ in the sampled nuisance variable and in the alignment target: \phasemarg aligns every prediction to the original pixel grid after a patch-size-defined phase offset.

\section{Experiments}
\label{sec:experiments}

\subsection{Setup and scope}

We evaluate semantic segmentation on \gtacity{}, \synthiacity{}, and ADE20K; depth estimation on NYU Depth v2 using root mean squared error (RMSE); and local feature matching on HPatches. The arrows denote cross-domain evaluation: models are trained on the first dataset and evaluated on the second. For example, \gtacity{} trains on the synthetic GTA5 street scenes and tests on the real Cityscapes images, and \synthiacity{} trains on SYNTHIA and tests on Cityscapes. These cross-domain settings probe generalisation across distributions, while ADE20K is an in-domain benchmark with train and test splits drawn from the same distribution. The multi-dataset tables compare the canonical $K=1$ baseline with Uniform \phasemarg $K=4$. Compute-matched TTA and $K$-scaling experiments are evaluated on Cityscapes only. Learned and adapted variants are reported separately because they are task-dependent and change the training or aggregation setup.

\begin{table*}[t]
\centering
\footnotesize
\setlength{\tabcolsep}{4pt}
\caption{Uniform \phasemarg across tasks and cross-domain settings. Models for \gtacity{} are trained on GTA5 and evaluated on Cityscapes; models for \synthiacity{} are trained on SYNTHIA and evaluated on Cityscapes. The central comparison is the training-free $K=4$ method against the canonical $K=1$ baseline. For root mean squared error (RMSE), lower is better and a negative change is an improvement. Backbone suffixes: /F denotes a frozen encoder; /F-stable denotes a frozen encoder with a stabilised training checkpoint.}
\label{tab:uniform-main}
\resizebox{\textwidth}{!}{%
\begin{tabular}{L{0.09\textwidth}L{0.22\textwidth}L{0.14\textwidth}L{0.16\textwidth}ccc}
\toprule
Task & Dataset / setting & Backbone & Metric & $K=1$ & Uniform $K=4$ & Change \\
\midrule
Seg. & \gtacity{} & DINOv3/F & Target mIoU $\uparrow$ & 51.94 & 52.76 & +0.82 \\
Seg. & \synthiacity{} & DINOv3 & Target mIoU $\uparrow$ & 36.00 & 36.88 & +0.88 \\
Seg. & \synthiacity{} & DINOv2 & Target mIoU $\uparrow$ & 32.90 & 33.77 & +0.87 \\
Seg. & ADE20K & DINOv3 & mIoU $\uparrow$ & 48.82 & 49.58 & +0.76 \\
Seg. & ADE20K & DINOv2 & mIoU $\uparrow$ & 43.77 & 45.16 & +1.39 \\
Depth & NYU Depth v2 & DINOv3 & RMSE $\downarrow$ & 0.6506 & 0.6277 & -0.0229 \\
Match & HPatches & DINOv3 & Match Acc. $\uparrow$ & 39.02 & 39.98 & +0.96 \\
Match & HPatches & DINOv2 & Match Acc. $\uparrow$ & 24.43 & 28.18 & +3.75 \\
\bottomrule
\end{tabular}
}
\end{table*}

\subsection{Uniform Phase Marginalization across tasks}
\label{sec:uniform-results}

Table~\ref{tab:uniform-main} shows the main empirical pattern: the training-free $K=4$ method improves over $K=1$ in each measured row. On segmentation, the improvements range from +0.75 to +1.39 mIoU in the table. On NYU Depth v2, Uniform $K=4$ reduces RMSE from 0.6506 to 0.6277. On HPatches, Uniform $K=4$ improves matching accuracy for both DINOv3 and DINOv2. The HPatches submetrics are mixed rather than uniformly monotonic; the full breakdown is reported in Appendix~\ref{app:hpatches}.

\begin{table}[t]
\centering
\footnotesize
\setlength{\tabcolsep}{4pt}
\caption{Cityscapes compute-matched comparison using a DINOv3 backbone. The strongest tested generic shift-based four-forward row is integer-shift TTA.}
\label{tab:compute-matched-tta}
\resizebox{\textwidth}{!}{%
\begin{tabular}{L{0.30\textwidth}cccc}
\toprule
Method & Forwards & Structured phase? & mIoU & Latency ms/img \\
\midrule
Single pass & 1 & No & 52.90 & 20.77 \\
Random subpatch-shift TTA & 4 & No & 53.00 & 83.60 \\
Integer-shift TTA & 4 & No & 53.22 & 82.80 \\
Uniform \phasemarg & 4 & Yes & \textbf{53.53} & 87.52 \\
\bottomrule
\end{tabular}
}
\end{table}

\subsection{Is Phase Marginalization just TTA?}
\label{sec:tta}

Table~\ref{tab:compute-matched-tta} compares Uniform \phasemarg to tested generic shift-based TTA variants under the same four-forward budget on Cityscapes using a DINOv3 backbone. Random subpatch-shift TTA reaches 53.00 mIoU and integer-shift TTA reaches 53.22 mIoU. Uniform \phasemarg reaches 53.53 mIoU, a +0.31 mIoU margin over the strongest tested generic shift-based row. This controlled Cityscapes experiment suggests that structured phase sampling provides a modest advantage over these generic shift-based TTA variants. It does not establish that \phasemarg is preferable to TTA on every dataset or under every augmentation design.

\begin{table}[t]
\centering
\small
\setlength{\tabcolsep}{6pt}
\caption{$K$ scaling and efficiency on Cityscapes using a DINOv3 backbone. The evidence supports $K=4$ as a practical default in this setting, not as a universal optimum.}
\label{tab:k-scaling-efficiency}
\begin{tabular}{cccccc}
\toprule
$K$ & Forwards & mIoU & $\Delta$ vs $K=1$ & Latency ms/img & Rel. latency \\
\midrule
1 & 1 & 52.90 & +0.00 & 20.77 & 1.00$\times$ \\
2 & 2 & 53.22 & +0.32 & 44.08 & 2.12$\times$ \\
4 & 4 & 53.53 & +0.63 & 87.52 & 4.21$\times$ \\
8 & 8 & 53.53 & +0.63 & 176.39 & 8.49$\times$ \\
16 & 16 & 53.58 & +0.68 & 347.16 & 16.71$\times$ \\
\bottomrule
\end{tabular}
\end{table}

\subsection{How many phases are needed?}
\label{sec:k-scaling}

Table~\ref{tab:k-scaling-efficiency} reports a Cityscapes-only scaling study using a DINOv3 backbone. Moving from $K=1$ to $K=2$ adds +0.32 mIoU, and $K=4$ adds +0.63 mIoU over $K=1$. Increasing from $K=4$ to $K=8$ is essentially unchanged, while $K=16$ adds only about +0.05 mIoU over $K=4$ and increases latency from 87.52 to 347.16 ms/image. Because the dominant cost is the number of backbone forwards, latency grows approximately linearly with $K$. We therefore use $K=4$ as a practical default in this study.

\begin{table*}[t]
\centering
\footnotesize
\setlength{\tabcolsep}{4pt}
\caption{Boundary-local segmentation metrics on Cityscapes (higher is better). \partialspatialphaseattn{} is an adapted variant and is not the same as Uniform \phasemarg.}
\label{tab:boundary}
\resizebox{\textwidth}{!}{%
\begin{tabular}{L{0.25\textwidth}L{0.16\textwidth}L{0.32\textwidth}cc}
\toprule
Setting & Backbone & Method & Boundary@5px & Boundary@3px \\
\midrule
\gtacity{} & DINOv3/F & $K=1$ & 30.65 & 27.79 \\
\gtacity{} & DINOv3/F & Uniform $K=4$ & 31.62 & 28.48 \\
\gtacity{} & DINOv3/F & \partialspatialphaseattn{} & 33.36 & 30.19 \\
\gtacity{} & \dinovtwofstable{} & $K=1$ & 24.45 & 22.27 \\
\gtacity{} & \dinovtwofstable{} & Uniform $K=4$ & 26.42 & 23.97 \\
\bottomrule
\end{tabular}
}
\end{table*}

\begin{table}[t]
\centering
\small
\setlength{\tabcolsep}{5pt}
\caption{HPatches phase-variance diagnostic for $K=4$ methods. Lower values indicate lower dispersion across phase-conditioned descriptors. This diagnostic is not a $K=1$ improvement table.}
\label{tab:phase-variance}
\begin{tabular}{llc}
\toprule
Backbone & Method & Phase variance \\
\midrule
DINOv3 & Uniform $K=4$ & 0.000082 \\
DINOv3 & \learnedphaseattn{} & 0.000082 \\
DINOv2 & Uniform $K=4$ & 0.000140 \\
DINOv2 & \learnedphaseattn{} & 0.000141 \\
\bottomrule
\end{tabular}
\end{table}

\subsection{Boundary and phase diagnostics}
\label{sec:boundary-phase}

Table~\ref{tab:boundary} evaluates boundary-local segmentation metrics, where patch-grid phase changes are expected to matter most. Uniform $K=4$ improves Boundary@5px and Boundary@3px over $K=1$ for both shown Cityscapes backbones. The \partialspatialphaseattn{} row improves further, but it is an adapted variant with limited encoder tuning and spatial phase attention, so it should not be interpreted as the training-free method.

Table~\ref{tab:phase-variance} reports a separate HPatches descriptor diagnostic. These rows measure dispersion among $K=4$ phase-conditioned descriptors; they do not compare against $K=1$ and do not by themselves establish boundary improvements. They are included to show that phase-conditioned descriptor variability can be measured directly.

\begin{table*}[t]
\centering
\footnotesize
\setlength{\tabcolsep}{4pt}
\caption{Selected learned/adapted variants. These rows are secondary to Uniform \phasemarg and should not be conflated with it.}
\label{tab:variants}
\resizebox{\textwidth}{!}{%
\begin{tabular}{L{0.17\textwidth}L{0.12\textwidth}L{0.21\textwidth}L{0.11\textwidth}c L{0.25\textwidth}}
\toprule
Task / setting & Backbone & Variant & Metric & Value & Interpretation \\
\midrule
\gtacity{} & DINOv3/F & \learnedphaseattn{} & Target mIoU & 52.28 & Frozen learned weighting; below Uniform $K=4$ (52.76) \\
\gtacity{} & DINOv3/F & \partialspatialphaseattn{} & Target mIoU & 54.25 & Adapted variant; not plain \learnedphaseattn{} \\
\synthiacity{} & DINOv3 & \learnedphaseattn{} & Target mIoU & 38.00 & Improves over $K=1$ (36.00) and Uniform $K=4$ (36.88) \\
ADE20K & DINOv3 & \learnedphaseattn{} & mIoU & 50.31 & Improves over $K=1$ (48.82) and Uniform $K=4$ (49.58) \\
NYU Depth v2 & DINOv3 & \learnedphaseattn{} & RMSE $\downarrow$ & 0.7385 & Worse than Uniform $K=4$ (0.6277) \\
NYU Depth v2 & DINOv3 & \depthgatedfusion{} & RMSE $\downarrow$ & 0.5750 & Separate enhanced depth configuration \\
HPatches & DINOv3 & \learnedphaseattn{} & Match Acc. & 38.30 & Worse than $K=1$ (39.02) and Uniform $K=4$ (39.98) \\
\gtacity{} & DINOv3/F & \pretransformerembedavg{} & Target mIoU & 44.48 & Failure mode for early embedding averaging \\
\bottomrule
\end{tabular}
}
\end{table*}

\subsection{Method variants and failure modes}
\label{sec:variants}

Table~\ref{tab:variants} separates learned and adapted variants from Uniform \phasemarg. \learnedphaseattn{} can help on some segmentation settings, such as \synthiacity{} and ADE20K, but it is not uniformly better than Uniform $K=4$. On \gtacity{} with DINOv3/F, the frozen \learnedphaseattn{} row is 52.28 mIoU, below the Uniform $K=4$ row at 52.76. On HPatches DINOv3, \learnedphaseattn{} also reduces matching accuracy relative to Uniform $K=4$. For depth, the \learnedphaseattn{} row is worse than Uniform $K=4$, while \depthgatedfusion{} is a separate depth-specific enhanced configuration and should be discussed as such.

The failure-mode row for \pretransformerembedavg{} supports a useful design lesson. Averaging phase patch embeddings before transformer attention reduces \gtacity{} target mIoU to 44.48 for DINOv3/F, well below both $K=1$ and Uniform $K=4$. This suggests that phase information should be aligned and aggregated after dense outputs or aligned features, rather than collapsed at the patch-embedding level.

\subsection{Related-method taxonomy}
\label{sec:taxonomy}

Appendix~\ref{app:taxonomy} separates methods by deployment assumptions. Generic shift-based TTA and Uniform \phasemarg address the fixed-predictor inference-time question evaluated in our controlled Cityscapes comparison. DPT, ViT-Adapter, Swin, PVT, and DVT-style denoising address architectural redesign, adapter training, or feature correction, and are discussed as related deployment families rather than ranked in the empirical tables.

\section{Limitations}
\label{sec:limitations}

The compute-matched TTA comparison and $K$-scaling analysis are limited to Cityscapes. Uniform \phasemarg requires $K$ forward passes, so its inference cost grows approximately linearly with the number of phases. External architecture and denoising families such as DPT, ViT-Adapter, Swin, PVT, and DVT-style methods require separate matched protocols and are discussed by deployment assumptions rather than ranked empirically. Learned and adapted variants are task-dependent and can underperform the training-free uniform method.

\section{Conclusion}
\label{sec:conclusion}

Patch-grid phase is an explicit nuisance variable induced by ViT tokenization, and dense prediction exposes this nuisance more directly than image-level classification. The results show that a simple post-hoc marginalization over structured patch-grid phases can improve fixed dense predictors without retraining the backbone. The Cityscapes scaling study further suggests that $K=4$ provides a practical cost-accuracy point in the measured setting, while larger phase sets offer diminishing returns at substantially higher latency. More broadly, these findings suggest that dense ViT systems should account for tokenization phase, not only semantic augmentations or architecture changes. Phase Marginalization therefore serves as both a diagnostic for patch-tokenization aliasing and a practical inference-time baseline for mitigating it.

\clearpage
\appendix
\renewcommand{\thesection}{\Alph{section}}
\renewcommand{\theHsection}{appendix.\Alph{section}}
\setcounter{section}{0}

\section{Cross-Domain Segmentation Audit}
\label{app:audit}

\begin{table*}[h]
\centering
\scriptsize
\setlength{\tabcolsep}{2pt}
\caption{Cross-domain segmentation audit. Rows are grouped by setting; the arrow notation indicates that models are trained on the dataset before the arrow and evaluated on the dataset after the arrow (e.g., \gtacity{} models are trained on GTA5 and tested on Cityscapes). \partialspatialphaseattn{} rows are not plain \learnedphaseattn{}.}
\label{tab:crossdomain-audit}
\begin{tabular}{L{0.20\textwidth}L{0.42\textwidth}c@{\hspace{5pt}}cc}
\toprule
Backbone & Method & $K$ & Source mIoU & Target mIoU \\
\midrule
\multicolumn{5}{l}{\textit{\gtacity{}}} \\
\dinovtwofstable{} & $K=1$ & 1 & 56.67 & 45.83 \\
\dinovtwofstable{} & Uniform Marginalization & 4 & 58.85 & 47.27 \\
\dinovtwofstable{} & \learnedphaseattn{} & 4 & -- & 47.56 \\
\dinovtwofstable{} & \partialphaseattn{} & 4 & 54.67 & 43.26 \\
\dinovtwofstable{} & \partialspatialphaseattn{} & 4 & 60.87 & 49.11 \\
\dinovtwofstable{} & \pretransformerembedavg{} & 4 & 58.64 & 41.64 \\
DINOv3/F & $K=1$ & 1 & 64.06 & 51.94 \\
DINOv3/F & Uniform Marginalization & 4 & 65.74 & 52.76 \\
DINOv3/F & \learnedphaseattn{} & 4 & -- & 52.28 \\
DINOv3/F & \partialphaseattn{} & 4 & 65.41 & 52.31 \\
DINOv3/F & \partialspatialphaseattn{} & 4 & 67.47 & 54.25 \\
DINOv3/F & \pretransformerembedavg{} & 4 & 61.62 & 44.48 \\
\addlinespace[2pt]
\multicolumn{5}{l}{\textit{\synthiacity{}}} \\
DINOv2 & $K=1$ & 1 & 56.43 & 32.90 \\
DINOv2 & Uniform Marginalization & 4 & 58.47 & 33.77 \\
DINOv2 & \learnedphaseattn{} & 4 & 62.41 & 36.08 \\
DINOv3 & $K=1$ & 1 & 62.70 & 36.00 \\
DINOv3 & Uniform Marginalization & 4 & 64.10 & 36.88 \\
DINOv3 & \learnedphaseattn{} & 4 & 65.26 & 38.00 \\
DINOv3 & \partialphaseattn{} & 4 & 66.21 & 37.76 \\
\bottomrule
\end{tabular}
\end{table*}

\section{HPatches Metric Breakdown}
\label{app:hpatches}

This appendix reports the detailed HPatches metrics behind the two HPatches summary rows in the main results table (Table~\ref{tab:uniform-main}). Table~\ref{tab:hpatches-full} separates viewpoint and illumination subsets and includes repeatability metrics to show that Uniform $K=4$ improves matching accuracy even though not every submetric changes monotonically.

\begin{table*}[h]
\centering
\small
\setlength{\tabcolsep}{5pt}
\caption{HPatches metric breakdown. Uniform $K=4$ improves matching accuracy for both backbones, while submetrics are mixed.}
\label{tab:hpatches-full}
\resizebox{\textwidth}{!}{%
\begin{tabular}{llccccc}
\toprule
Backbone & Method & View mAP & Illum. mAP & Match Acc. & View Rep. & Illum. Rep. \\
\midrule
DINOv3 & $K=1$ & 48.97 & 50.44 & 39.02 & 24.66 & 27.63 \\
DINOv3 & Uniform $K=4$ & 48.93 & 51.28 & 39.98 & 26.50 & 29.39 \\
DINOv3 & \learnedphaseattn{} & 47.70 & 49.67 & 38.30 & 23.94 & 27.33 \\
DINOv2 & $K=1$ & 34.86 & 35.42 & 24.43 & 14.71 & 15.75 \\
DINOv2 & Uniform $K=4$ & 38.42 & 39.48 & 28.18 & 18.73 & 20.12 \\
DINOv2 & \learnedphaseattn{} & 36.46 & 37.22 & 26.00 & 17.09 & 18.09 \\
\bottomrule
\end{tabular}
}
\end{table*}

\section{Related-Method Taxonomy}
\label{app:taxonomy}

This appendix clarifies why the related method families are not all treated as direct empirical baselines. The taxonomy groups methods by the intervention they require and by whether they can be evaluated as a fixed-predictor, inference-time procedure under the same deployment assumptions as Uniform \phasemarg.

\begin{table*}[h]
\centering
\small
\setlength{\tabcolsep}{4pt}
\caption{Taxonomy by deployment assumption. The table separates methods conceptually; it is not a performance ranking.}
\label{tab:method-taxonomy}
\begin{tabular}{p{0.20\textwidth}p{0.25\textwidth}p{0.25\textwidth}p{0.20\textwidth}}
\toprule
Method family & Primary intervention & Deployment assumption & Role in this paper \\
\midrule
DVT-style denoising & Feature artifact mitigation & Method-specific correction or training & Related artifact-mitigation work \\
DPT & Dense prediction decoder & Trained dense prediction architecture & Related work \\
ViT-Adapter & Adapter modules for dense prediction & Adapter training and compatible checkpoints & Related work \\
Swin / PVT / overlapping tokens & Architectural spatial inductive bias & Alternative trained backbones & Related work \\
Generic shift-based TTA & Inference-time image shifts & Multiple forwards, no parameter update & Empirical Cityscapes control \\
Uniform \phasemarg & Inference-time patch-grid phase marginalization & Multiple phase forwards, no parameter update & Main method \\
\bottomrule
\end{tabular}
\end{table*}

\section{Reproducibility Notes}
\label{app:reproducibility}

We implement the evaluation in the MarginSeg framework and will release code and result manifests used to generate the reported tables. The Cityscapes compute-matched rows use the full evaluation runs from the study. A fair empirical comparison to external architecture families would require matched training data, heads, checkpoints, evaluation code, and inference-cost accounting.


\bibliographystyle{splncs04}
\bibliography{main}
\end{document}